%% file: sample-acmtog.tex
\documentclass[acmtog, authorversion]{acmart}
\usepackage{float} 
\usepackage{booktabs} 
\usepackage{subfig}
\usepackage{subfloat}

\setcitestyle{square}

\usepackage[ruled]{algorithm2e} 

\SetAlFnt{\small}
\SetAlCapFnt{\small}
\SetAlCapNameFnt{\small}
\SetAlCapHSkip{0pt}
\IncMargin{-\parindent}

\begin{document}
\title{CIKM AnalytiCup 2017 -- Lazada Product Title Quality Challenge: An Ensemble of Deep and Shallow Learning to predict the Quality of Product Titles} 

\author{Karamjit Singh}
\affiliation{%
  \institution{TCS Research}
  \city{Delhi}
  \country{India}
}
\author{Vishal Sunder} 
\affiliation{%
  \institution{TCS Research}
  \city{Delhi}
  \country{India}
}

\renewcommand\shortauthors{Singh, K. et al}

\begin{abstract}
We present an approach where two different models (\textit{Deep} and \textit{Shallow}) are trained separately on the data and a weighted average of the outputs is taken as the final result. For the Deep approach, we use different combinations of models like Convolution Neural Network, pretrained word2vec embeddings and LSTMs to get representations which are then used to train a Deep Neural Network. For Clarity prediction, we also use an \textit{Attentive Pooling} approach for the \textit{pooling} operation so as to be aware of the \textit{Title}-\textit{Category} pair. For the shallow approach, we use boosting technique LightGBM on features generated using title and categories. We find that an ensemble of these approaches does a better job than using them alone suggesting that the results of the deep and shallow approach are highly complementary.
\end{abstract}

\maketitle

\input{samplebody-journals}

\end{document}

%% file: samplebody-journals.tex
\section{Introduction}
\label{intro}
In this data challenge, we were given a set of product attributes like title, sub-categories, the country where the product is marketed etc. Given these attributes, the task was twofold. First, to predict whether the product title is 'Clear' and second whether it is 'Concise'. The provided training data contained 36283 samples which were manually labelled by Lazada's internal QC team under set guidelines.

We treat this as a binary classification problem which we try to solve separately using a \textit{Deep} and a \textit{Shallow} model and finally take a weighted sum of their output probabilities as the result. 

For the \textit{Deep Approach}, we use deep models with Convolution Neural Networks (CNN) \citep{kim2014convolutional} and Long Short Term Memory (LSTM) \cite{hochreiter1997long} through which we get deep representations for the \textit{Title}/\textit{Category}. These representations were learnt through input features which were engineered according to the task (\textit{Clarity}/\textit{Conciseness} prediction).. We also try a fairly new approach for the pooling layer i.e. \textit{Attentive Pooling} introduced in \citep{dos2016attentive} and achieve impressive results.

For the \textit{Shallow Approach}, we engineered various features based on counts, syntax matching, semantic matching, etc, using title and categories. Further, we use the LightGBM algorithm as a classification algorithm to predict the probabilities of clarity and conciseness of title. 

\section{Related Work}
\label{relw}
In recent years, Deep Learning approaches have found popularity for NLP tasks \citep{collobert2011natural}, \citep{collobert2008unified}. In particular, text classification has been achieved using Convolutional Neural Networks by \citep{kim2014convolutional}, who uses it for sentiment and question classification among other things. \cite{zhou2015c} have used a unified model of LSTM and CNN for text classification by using a CNN to extract high level phrase representation which are then fed to a LSTM for obtaining final representation. To get an improved representation, \citep{dos2016attentive} introduced an attentive pooling approach which is a two-way attention mechanism for discriminative model training.

\section{Feature engineering} \label{sec:feat}
In this section, we present the set of features divided into various classes:

\begin{itemize}
\item \textbf{Given features:} 1) Country(categorical), 2) \textit{Price:} normalize after taking log transformation, 3), Level(categorical), 4) Category-1(categorical), 5) Category-2(categorical), 6) Category-3(categorical).

\item \textbf{Title-Counts:} We generate features using title in two ways: a) After cleaning (AC): removing all non alphabatic and non-numeric characters, b) Before cleaning (BC): No character removed

7) \textit{Number of words(AC)}: calculate number of words using space character as split.

8) \textit{Max length(AC)}: maximum length (number of characters) of a word in a title.

9) \textit{Min length(AC)}: minimum length (number of characters) of a word in a title

10) \textit{Average length(AC)}: average length (number of characters) of all words in a title.

Similarly, 11) \textit{Number of words(BC)}, 12) \textit{Max length(BC)}, 13) \textit{Min length(BC)}, 14) \textit{Average length(BC)}

15) \textit{Contains digit}: 1 or 0 whether title contains digit or not

16) \textit{Non-Alpha Per}: percentage of non-alphabatic characters in a title

\item \textbf{Title String matching:} We calculate the jaro-winkler\citep{winkler1999state} distance between each pair of word in a title and generate following features:

17) \textit{Average of string distance:} An average distance of all pairs in a title.

18) \textit{Max String distance count:} Percentage of pairs having distance greater than 0.85.

19) \textit{Min String distance count:} Percentage of pairs having distance less than 0.1.

20) \textit{Full String match count:} Percentage of pairs having distance equal to 1.

21) \textit{Sum of String distance:} Sum of distances of all pairs

\item \textbf{Title Semantic matching:} Similar to string matching features, we also calculate semantic similarity based features. For each pair of words in a title, we calculate the cosine distance between vectors of words generated from Common crawl google glove~\cite{pennington2014glove}. Further, we normalize cosine distance between 0 to 1 and generate following features:

22) \textit{ Average of semantic distance}, 23) \textit{Max semantic distance count}, 24) \textit{Min semantic distance count}, and 25) \textit{Full String match count}.

26) \textit{Percentage of present:} percentage of words in a title which are present in glove dictionary.

27) \textit{Unique non-presenter:} percentage of unique words in a title among non-presenter in a glove dictionary.

\item \textbf{Title Category semantic matching:} We generate features which captures the semantic matching of title with each of the three categories. For a title for sample $i$, we generate a $AvgVecT_i$ by taking average of the vectors all words in a title. Similarly, we generate average vector $AvgVecC_i$ for corresponding category and then we take cosine distance between these two vectors

28) \textit{Average Sem distance Title-Cat1}, 29) \textit{Average Sem distance Title-Cat2}, 30) \textit{Average Sem distance Title-Cat3}

\item \textbf{Title Other categories:} We generate features which captures the semantic matching of a title with all categories states except its own. Process of generating these features is explained as follows:

Let $k+1$ be the total number of possible states of category-1. For a title of sample $i$, omitting the category of sample $i$, we prepare $k$ matrices $M^i_j$, $j =1,2,..., k$ of order $n_i*m_j$, where $n_i$ is number of words $\{t_1, t_2,...., t_{n_i}\}$ in a title of sample $i$ and $m_j$ is the number of words $\{c_1, c_2, ..., c_{m_j}\}$ in category-1 $j$. Each cell $(n_i, m_j)$ of matrix $M^i_j$ contains the cosine distance between the vectors (generated from glove) of words $t_{n_i}$ and $c_{m_j}$. Further, for each sample $i$, we generate a list $L_i$ containing $k$ numbers, where each number is an average of matrix $M^i_j$, $j=1,2,..., k$. Using list $L_i$, we generate following features for category-1 and category-2:

31) \textit{Sum of Title-Cat-1 Matrix} and 32) \textit{Sum of Title-Cat-2 Matrix}: Taking the sum of $L_i$

33) \textit{Max of Title-Cat-1 Matrix} and 34) \textit{Max of Title-Cat-2 Matrix}: Taking the max of $L_i$

35) \textit{Min of Title-Cat-1 Matrix} and 36) \textit{Min of Title-Cat-2 Matrix}: Taking the min of $L_i$.

\item \textbf{Title category Matrix:} We generate features capturing semantic matching of title and its corresponding category. For each title in a sample $i$, we prepare a matrix $M_i$ of order $n_i, m_i$, where $n_i$ is the number of words in a title $i$ and $m_i$ is the number of words in a corresponding category. Each cell of this matrix contains the the cosine distance between the vectors (generated from glove) of words of title and category. Using the matrix $M_i$, we generate following features for all three categories:

37) \textit{Max of Title-Cat1 matrix}, 38) \textit{Max of Title-Cat2 matrix}, 39) \textit{Max of Title-Cat3 matrix}: Taking max of matrix $M_i$

\item \textbf{Shallow Attention}: We apply the Attentive Pooling of attention (Section \ref{attnp}) to the \textit{P2M} representation (Section \ref{p2m}) of the title and its sub-categories one by one which gives us $3$ pairs of vectors. The $cosine$ distance for each pair is computed and this gives us $3$ features 40-42) $v_1, v_2, v_3$. Finally, we apply this attention to the \textit{P2M} representation of the title and a single category obtained by merging the three sub-categories together. This gives us another pair of vectors. Taking the cosine distance of this pair gives us a fourth feature 43) $v_4$.

\item \textbf{Others:}

44) \textit{Sexy:} Whether the title contains a word `sexy' or not

45) \textit{Percentage of capital letters in a title}

\end{itemize}

\section{Methodology}
\label{sec:method}
The given training set contained $36283$ samples. $80\%$ of this set was used for training the model while the remaining $20\%$ was used as a $Holdout$ set for validation.
\subsection{Clarity}

\subsubsection{Deep Approach-CLR}
\begin{figure}[H]
\centering
\includegraphics[width=0.5\textwidth,height=3cm]{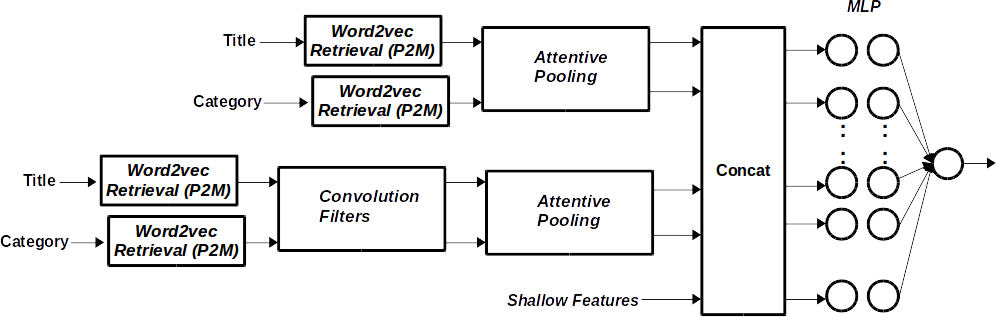}
\caption{Deep model for Clarity prediction.}
\label{fig:deep_clar}
\end{figure}

An overview of the deep model is given in Figure \ref{fig:deep_clar}. 

The following methods are used to represent a Title or Category as a dense matrix.

\subsubsection*{Phrase2Mat (P2M)}
\label{p2m}
First, we join the three sub-categories so as to form a single category phrase. Then tokenization is done on a phrase (title or category), removing all out of vocabulary words, and appending the pretrained $word2vec$ ($GoogleNews$) representation \citep{mikolov2014word2vec} obtained from of the tokens to form the \textit{Phrase2Mat} matrix.

Hence, we get $\mathbf{T}_{w2v}$ $\in$ $\mathbb{R}^{n \times M}$ for the Title and $\mathbf{C}_{w2v}$ $\in$ $\mathbb{R}^{n \times N}$ for the Category where $M$ and $N$ are lengths of title and category respectively.
\subsubsection*{Convolutional Neural Network (CNN)}
We also use a $CNN$ to get another set of dense representation for a phrase \cite{kim2014convolutional}. In particular, a $filter$ $\mathbf{W}$ $\in$ $\mathbb{R}^{n \times h}$ with \textit{tanh} activation is applied to a window of $h$ words to produce a feature. Thus, for any matrix $\mathbf{X} = [\mathbf{x}_1,\mathbf{x}_2,...,\mathbf{x}_N]$ where $\mathbf{x}_{i}$ $\in$ $\mathbb{R}^{n}$, applying the filter $\mathbf{W}$ with a \textit{stride} of 1 gives us $N-h+1$ features. Applying $F$ such filters on $\mathbf{T}_{w2v}$ and $\mathbf{C}_{w2v}$ defined above gives \textit{feature maps} $\mathbf{T}_{cnn}$ and $\mathbf{C}_{cnn}$ $\in$ $\mathbb{R}^{F \times (N-h+1)}$

To get vectorized representation from matrices, various pooling methods may be used (such as $AveragePooling$ and $MaxPooling$). We use $AttentivePooling$ method which is much more sophisticated.

\subsubsection*{Attentive Pooling (AttnP)}
\label{attnp}
Attentive pooling \citep{dos2016attentive} is a way of pooling which allows the pooling operation to be aware of the input pair, in a way that information from the category can directly influence the computation of the title representation $\mathbf{r}^{T}$, and vice versa. Thus, given a $(Title,Category)$ matrix pair $(\mathbf{T},\mathbf{C})$, our aim is to find corresponding vector representation $(\mathbf{r}^T,\mathbf{r}^C)$.

Given input matrices $\mathbf{T}$ and $\mathbf{C}$, we try to find a \textit{soft alignment} between $\mathbf{T}$ ($Title$) and $\mathbf{C}$ ($Category$) after which a weighted average pooling is applied.

A detailed explanation of this approach can be found in \citep{dos2016attentive}.

Thus, from attentive pooling, we obtain $(\mathbf{r}^{T_{w2v}},\mathbf{r}^{C_{w2v}})$ and $(\mathbf{r}^{T_{cnn}},$\newline$\mathbf{r}^{C_{cnn}})$ from $(\mathbf{T}_{w2v},\mathbf{C}_{w2v})$ and $(\mathbf{T}_{cnn},\mathbf{C}_{cnn})$ respectively.

The final representation $\mathbf{r}^{final}$ $\in$ $\mathbb{R}^{2n+2F+sh}$ is obtained by stacking the representation obtained from the deep network with the engineered features from the \textit{Shallow Approach}. Here $sh$ is the number of engineered features.

This final representation $\mathbf{r}^{final}$ is given as input to a fully connected Neural Network with $ReLu$ activation in the hidden layers and two neurons with $softmax$ activation as final output. We use the $Adam$ optimizer to train this fully connected network by minimizing the cross-entropy loss. We also used $dropout$ for regularization.

The various hyperparameter setting for this model is given in Table \ref{tab:one} of Appendix \ref{hypset}.

\subsubsection{Shallow Approach-CLR} For clarity, we use 24 features given as: Given features (1 to 6), Title Counts (7 to 11), and 15, 16, Title Category semantic matching (28 to 30), Title category Matrix(37 to 39), Other features(40), and Deep features (42 to 45) and we use LightGBM package~\footnote{https://github.com/Microsoft/LightGBM} proposed by~\cite{meng2016communication} as a classification algorithm.

\subsection{Conciseness}
\subsubsection{Deep Approach-CON}
\begin{figure}[H]
\centering
\includegraphics[width=0.5\textwidth,height=3cm]{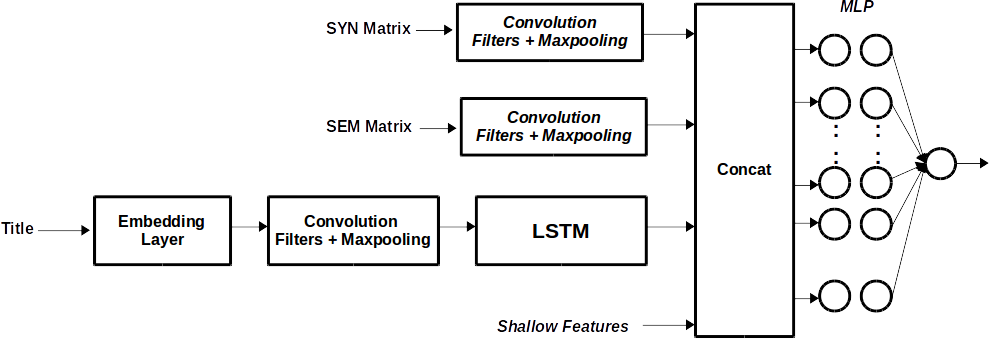}
\caption{Deep model for Conciseness prediction.}
\label{fig:deep_cons}
\end{figure}

An overview of the deep model is given in Figure \ref{fig:deep_cons}

\subsubsection*{Intra-Title Features (ITF)}
To predict Conciseness, we utilised the features contained in the Title only and use features which capture Intra-Title relations. To do this, we extract the semantic and syntactic relationships between tokens in a Title by generating two matrices $\mathbf{SEM}$ and $\mathbf{SYN}$ respectively.
\begin{description}
\item[Semantic Relation]
We tokenize the Title and compute the $cosine$ distance between the $word2vec$ representation of all possible pair of tokens. This gives us a symmetric matrix $\mathbf{SEM}$ $\in$ $\mathbb{R}^{N \times N}$ where $N$ is the maximum length of a Title.
\item[Syntactic Relation]
We tokenize the Title and compute the \textit{Jaro-Winkler} distance \citep{winkler1999state} between all possible pair of tokens. This gives us a symmetric matrix $\mathbf{SYN}$ $\in$ $\mathbb{R}^{N \times N}$ where $N$ is the maximum length of a Title.
\end{description}
We remove all out of vocabulary words in generating both the above matrices.

For computing the matrices, the intuition is that, given a Title, we need to know how related/unrelated are the tokens in it. A non-concise title will typically have more noisy tokens or tokens that don't have much relation with the rest of the title. Hence, we propose that looking at the pairwise relations (semantic/syntactic) between tokens can give an overall picture of the coherence of a title. This relation is precisely represented by the \textbf{SEM} and \textbf{SYN} matrix.

A dense representation is obtained from these relation matrices by using a \textit{CNN} on them. A convolution with a filter $\mathbf{W}$ $\in$ $\mathbb{R}^{N \times h}$ with \textit{tanh} activation is applied to both $\mathbf{SEM}$ and $\mathbf{SYN}$ with a stride 1 to give feature maps $\mathbf{F}_{SEM}$ and $\mathbf{F}_{SYN}$ $\in$ $\mathbb{R}^{F \times (N-h+1)}$.

We then apply a max-over-time pooling operation \citep{collobert2008unified} over each row of the feature map to obtain final representations $\mathbf{r}^{SEM}$ and $\mathbf{r}^{SYN}$ $\in$ $\mathbb{R}^F$ respectively.

\subsubsection*{CNN-LSTM on Title}
We also use a CNN and an LSTM \citep{zhou2015c} together, directly on the title to get a deep representation of the same.
A title of length $N$ is passed through an embedding layer to get a matrix $\mathbf{X}_{title}$ $\in$ $\mathbb{R}^{n \times N}$ where $n$ is the embedding dimension. These embeddings are initialized to \textit{word2vec-GoogleNews} and fine-tuned during training (We also experimented without tuning the embeddings and Table \ref{DLresults} (b) contains results for both).

A CNN filter $\mathbf{W}$ $\in$ $\mathbb{R}^{n \times h}$ with \textit{tanh} activation is applied to $\mathbf{X}_{title}$ with stride 1 to obtain a feature map $\mathbf{F}_{title}$ $\in$ $\mathbb{R}^{F \times (N-h+1)}$. To this feature map, a max-over-time-pooling is applied with a \textit{pool size} of 2 to obtain a matrix $\mathbf{Y}$ $\in$ $\mathbb{R}^{F \times (N-h+1)/2}$ which serves as an input to the LSTM \citep{hochreiter1997long}. The LSTM has F timesteps. The output of the last timestep of the LSTM is $\mathbf{r}^{CNN-LSTM}$ $\in$ $\mathbb{R}^{hl}$ where $hl$ is the number of hidden units in the LSTM. 

The final representation $\mathbf{r}^{final}$ $\in$ $\mathbb{R}^{hl+2F+sh}$ is obtained by stacking $\mathbf{r}^{SEM}$, $\mathbf{r}^{SYN}$, $\mathbf{r}^{CNN-LSTM}$ and the engineered features from the \textit{Shallow Approach}.
Here $sh$ is the number of engineered features. This final representation $\mathbf{r}^{final}$ is given as input to a fully connected Neural Network with $ReLu$ activation in the hidden layers and a single neuron with $sigmoid$ activation as final output. We use the $Adam$ optimizer to train this fully connected network by minimizing the cross-entropy loss. We also used $dropout$ for regularization.

The various hyperparameter setting for this model is given in Table \ref{tab:one} of Appendix \ref{hypset}.

\subsubsection{Shallow Approach-CON} For conciseness, we use all the features mentioned in section~\ref{sec:feat} except Title category Matrix features and use LightGBM algorithm as a classification algorithm.

\section{Lessons Learnt}
\label{lessons}
One of the key observations is that a simple \textit{weighted average} of the output of the \textit{Shallow} and \textit{Deep} approaches give better results than the two alone suggesting that the results of the two are highly complementary. However, \textit{Stacking}, which is a common method of learning a good ensemble of models did not give good results. This suggests that although the features from the deep and shallow approaches capture complementary information, we need to look into ways of ensemble them properly.

In particular, we observed that Shallow Learning does better in places where the Title is either too short or when a title comprises of too many numerals and proper nouns which clearly is a problem of out of vocabulary words for Deep Learning. 

In contrast, Deep Learning does better than Shallow Learning when the Title is quite long with unnecessary words and hence non-concise. On similar grounds, if a title is long, has fewer proper nouns and is concise/clear, Deep Learning does a better job in classifying them correctly.

We observe that predicting the clarity of a title more sensitive as compared to predicting conciseness because adding/removing even small number features effects highly on clarity score. Hence, in the shallow model as well as deep model, we use less number features for clarity as compared to conciseness.

\section{Analysis}
\label{sec:analysis}
We tried different combinations of the approaches described in \ref{sec:method}. Due to the fact that some approaches were tried on the validation set while others on the test set, it is difficult to compare them. Hence, the results reported in this section for the \textit{Deep} and \textit{Shallow approaches} are on the \textit{Holdout set}. The final best result is also shown on the provided test set.

\begin{table}%
\subfloat[Clarity]
{
\begin{tabular}{ll}
  \toprule
  P2M + Average Pooling     & 0.2360\\
  P2M + AttnP  & 0.2318\\
  CNN + MaxPooling    & 0.2295\\
  CNN + AttnP  & 0.2293\\
  P2M + CNN + AttnP & 0.2272\\
  P2M + Average Pooling + Shallow Features & 0.2340\\
  P2M + AttnP + Shallow Features & 0.2304\\
  CNN + MaxPooling + Shallow Features   & 0.2295\\
  CNN + AttnP + Shallow Features  & 0.2277\\
  P2M + CNN + AttnP + Shallow Features & \textbf{0.2265}\\
  \bottomrule
\end{tabular}
}
\quad
\subfloat[Conciseness]
{
\begin{tabular}{ll}
  \toprule
  P2M + CNN + AttnP + Shallow Features (w/o tuning) & 0.3466\\
  P2M + CNN + AttnP + Shallow Features (tuning) &0.3471\\
  CNN on Title + Shallow Features (w/o tuning)  & 0.3433\\
  CNN on Title + Shallow Features (tuning)  & 0.3416\\
  ITF + Shallow Features (w/o tuning) & 0.3532\\
  ITF + Shallow Features (tuning) & 0.3531\\
  CNN-LSTM on Title + Shallow Features (w/o tuning) & 0.3430\\
  CNN-LSTM on Title + Shallow Features (tuning) & 0.3420\\
  ITF + CNN-LSTM + Shallow Features (tuning) & \textbf{0.3379}\\
  \bottomrule
\end{tabular}
}
\caption{Deep Learning Results on Holdout set in RMSE}
\label{DLresults}
\end{table}%

\begin{table}
\centering
\begin{tabular}{l|l|l|l|l}
\hline
\textbf{Methods:} & \multicolumn{2}{c|}{Clarity} & \multicolumn{2}{c}{Conciseness}\\  \hline
    & Holdout set & Test set
    & Holdout set & Test set  \\   \hline
    \textit{Deep Approach}     & 0.2265  & 0.2465  & 0.3379  & 0.3505\\
    \textit{Shallow Approach}  & 0.2271 & 0.2468  & 0.3295 & 0.3477 \\
    \textit{Ensemble}  & \textbf{0.2191} & \textbf{0.2438}  & \textbf{0.3187} & \textbf{0.3385}   \\ \hline

\end{tabular}
\caption{Final results on Test and Holdout Set in RMSE.}
\label{finalresult}
\end{table}

\subsection{Deep Approach}
The results for different combination of approaches tried for Clarity and conciseness are presented in Table \ref{DLresults}. The representation obtained from these ($\mathbf{r}^{final}$) are fed in to a Deep Neural Network as already explained in \ref{sec:method}.

We used a Quadro M3000M GPU machine (4 GB) for training which took around 3 minutes of training time for Clarity and 5.5 minutes of training time for Conciseness.

\subsection{Shallow Approach}

Table~\ref{tab:shallow} of Appendix \ref{hypset} shows the parameters used in LightGBM algorithm for both conciseness and clarity prediction. We also use different algorithms using same set of features. Table~\ref{tab:shallow-aglos} compares RMSE over holdout set using different algorithms. It shows that LightGBM outperform rest of the boosting and bagging techniques.

\begin{table}[H]
\centering
\begin{tabular}{l|l|l}
\hline
\textbf{Algorithms} & RMSE-CLR & RMSE-CON \\  \hline
 Random Forest & 0.2356 & 0.3465 \\
 GBM & 0.2305 & 0.3315 \\
 XGBoost & 0.2283 & 0.3305 \\
 LightGBM & \textbf{0.2271} & \textbf{0.3295} \\
\hline
\end{tabular}
\caption{Shallow Learning results on Holdout set in RMSE.}
\label{tab:shallow-aglos}
\end{table}

\subsection{Ensemble}
The final result on holdout and test set is shown in \ref{finalresult}. The results shown in this table are for the best performing models for both the Shallow and Deep approaches. In the weighted ensemble for clarity, we use  weights 0.55 and 0.45 for predicted probabilities from Deep and Shallow models respectively. For conciseness, we simply take the average (with weights 0.5) of the predicted probabilities.

\section{Conclusion} 
In this report, we have described the various approachs that we used for the "CIKM AnalytiCup 2017 -- Lazada Product Title Quality Challenge". As the final results prove, a weighted ensemble of Deep and Shallow Models outperform the individual approaches and hence set up a case for future work to learn a better ensemble of these models. 

\bibliographystyle{ACM-Reference-Format}
\bibliography{sample-bibliography}

\appendix
\section{Hyperparameter settings}
\label{hypset}
\begin{table}[H]
\centering
\begin{tabular}{l|l|l}
\hline
\textbf{Hyperparameter (Clar/Cons)} & Clarity & Conciseness \\  \hline
\textit{Word2Vec Embedding Dimension ($n$ / $n$)} & 300 & 300 \\
\textit{Maximum Title Length ($M$ / $N$)} & 45 & 45 \\
\textit{Maximum Category Length ($N$ / - )} & 10 & - \\
\textit{Filter Window Width ($h$ / $h$)} & 3 & 3 \\
\textit{Number of filters ($F$ / $F$)} & 100 & 128 \\
\textit{Dropout in CNN} & 0.2 & 0.2 \\
\textit{Dropout in LSTM (- / $hl$)} & - & 0.2 \\
\textit{Dropout in Embedding Layer} & - & 0.5 \\
\textit{Hidden units in LSTM} & - & 128 \\
\textit{Hidden Layers in Neural Net} & 5 & 3 \\
\textit{Hidden Layer dimension in Neural Net} & 10 & 50 \\
\textit{Learning Rate} & 0.0001 & 0.0001 \\
\hline
\end{tabular}
\caption{Hyperparameter setting for Deep Learning.}
\label{tab:one}
\end{table}

\begin{table}[H]
\centering
\begin{tabular}{l|l|l}
\hline
\textbf{Parameter} & Clarity & Conciseness \\  \hline
\textit{Learning Rate} & 0.03 & 0.03 \\
\textit{Colsample Bytree} & 0.6 & 0.65 \\
\textit{Max Depth} & 6 & -1 \\
\textit{Min Child Samples} & 10 & 10 \\
\textit{n Estimators} & 390 & 490 \\
\textit{Num Leaves} & 50 & 55 \\
\textit{Subsample} & 0.88 & 0.88 \\
\textit{Max Bin} & 255 & 255 \\
\hline
\end{tabular}
\caption{Hyperparameter setting for Shallow Learning.}
\label{tab:shallow}
\end{table}